\documentclass[10pt]{article}

\usepackage[preprint]{tmlr}

\usepackage[utf8]{inputenc} 
\usepackage{float}          
\usepackage{placeins}       
\usepackage[hyperfootnotes=false]{hyperref} 
\usepackage{url}            
\usepackage{booktabs}       
\usepackage{amsmath}        
\usepackage{amsfonts}       
\usepackage{graphicx}       
\usepackage{multirow}

\title{Beyond Static Structures: Linking Protein Dynamics to Function via DynaReason-Pro}

\author{%
  Zhaokang Liang\textsuperscript{1}\quad
  Mingze Yuan\textsuperscript{2}\quad
  Quanzheng Li\textsuperscript{3}\quad
  Wengong Jin\textsuperscript{1}\thanks{Corresponding author.}\\
  {\normalfont\small \textsuperscript{1}Northeastern University\quad
  \textsuperscript{2}Harvard University}\\
  {\normalfont\small \textsuperscript{3}Massachusetts General Hospital, Harvard Medical School}
}

\begin{document}

\maketitle

\begin{abstract}
Intrinsically disordered proteins (IDPs) carry out their functions through heterogeneous conformational ensembles that reflect their dynamics, yet most protein function prediction methods rely on sequence or a single static structure. To this end, we introduce DynaReason-Pro, the first framework to jointly predict GO terms from conformational ensembles and generate natural-language functional rationales. We evaluate DynaReason-Pro on a curated IDP benchmark from the Protein Ensemble Database (PED), with sequence- and structure-cluster-disjoint splits to prevent data leakage. Variability-stratified evaluation shows the clearest gains in the high-conformational-variability regime.\footnote{Large language model tools were used to assist with language editing and format checking.}
\end{abstract}

\section{Introduction}
Proteins are the molecular machines of life, carrying out the biochemical functions that underpin metabolism, signaling, and cellular organization. High-throughput sequencing has produced vast protein sequence data, with about 250 million entries in databases such as UniProt~\citep{uniprot2025uniprot}. Yet, functional characterization has not kept pace: fewer than 0.1\% of known proteins have experimentally validated annotations, owing to the high cost and limited throughput of wet-lab assays~\citep{wan2020protein}. This  annotation gap remains a central challenge in molecular biology and drug discovery, motivating the development of computational methods that can infer protein function at scale from sequence alone~\citep{friedberg2006automated,radivojac2013large}.

The prevailing paradigm assumes that amino-acid sequences determine protein structures~\citep{anfinsen1973principles}, which in turn determine function. Accordingly, most AI-based methods predict protein function either directly from sequence or from predicted static structures. While these approaches have made important progress, they share a fundamental limitation: they treat proteins as static objects. In reality, biological function is inherently dynamic~\citep{henzlerwildman2007dynamic}. Enzyme catalysis relies on transient conformational changes that position catalytic residues and stabilize transition states~\citep{henzlerwildman2007hierarchy}. Receptor signaling involves subtle shifts across energy landscapes~\citep{hilger2018structure}. Protein--protein interactions are mediated by flexibility and environmental context, not rigid lock-and-key fits~\citep{boehr2009role}. Static structures can suggest possible interactions, but they cannot explain how those interactions occur, change over time, or why they sometimes fail.

We introduce DynaReason-Pro, a unified framework for predicting protein function from conformational ensembles that encode dynamics-related structural heterogeneity. Its ensemble-conditioned ontology predictor uses residue-wise set attention to integrate structural states and autoregressively generate GO terms. An LLM-based reasoning module then combines these ontology anchors with the protein sequence, organism information, and InterPro annotations to generate functional descriptions and natural-language rationales. Together, these components connect structured ontology prediction with human-readable downstream functional interpretation.

We evaluate DynaReason-Pro on intrinsically disordered proteins (IDPs), whose heterogeneous ensembles provide a natural test of whether conformational information adds value beyond sequence or a single structure~\citep{tesei2024conformational}. We curate a benchmark from the Protein Ensemble Database (PED) with sequence- and structure-cluster-disjoint splits, and complement aggregate evaluation with conformational-variability stratification, controlled ablations, term-level case studies, and reasoning-trace visualizations. DynaReason-Pro's relative advantage is strongest in the high-conformational-variability regime.

\begin{figure}[t]
\centering
\includegraphics[width=\linewidth]{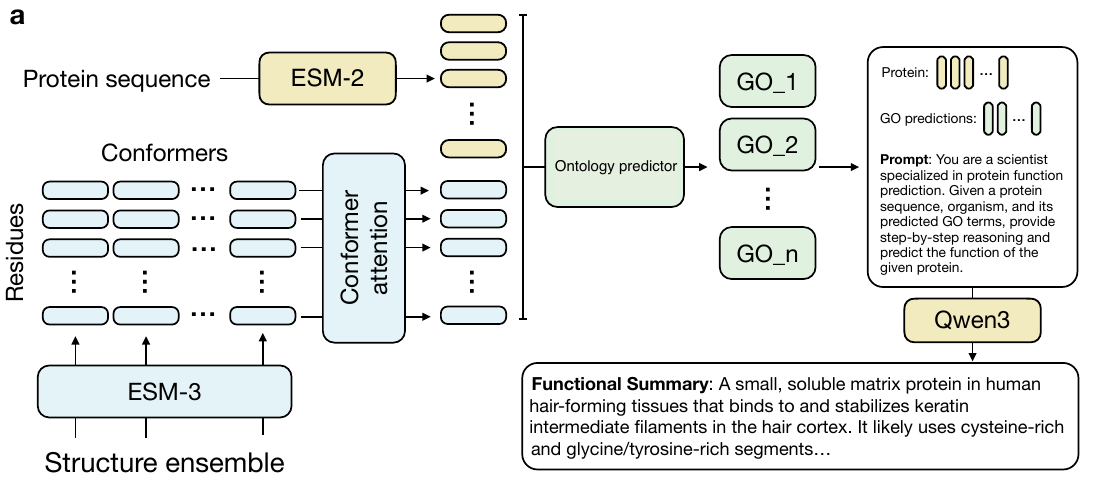}\\[-0.35em]
\includegraphics[width=0.48\linewidth]{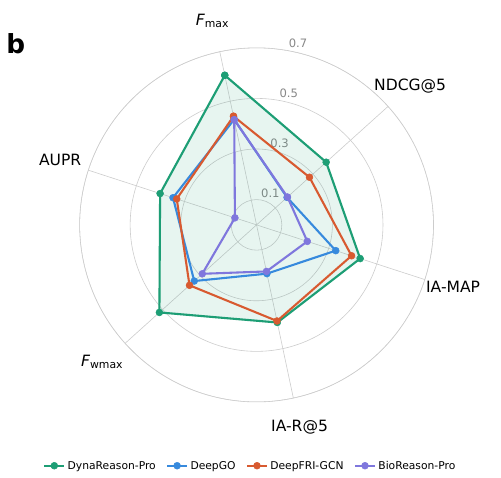}
\caption{\textbf{a}, Architecture of DynaReason-Pro. The ontology predictor integrates the protein sequence and residue-aligned conformational ensemble to generate GO terms; the LLM-based reasoning module then combines these ontology anchors with the protein sequence, organism information, and InterPro annotations to generate functional descriptions and natural-language rationales. \textbf{b}, Comparison with DeepGO, DeepFRI-GCN, and BioReason-Pro across \(F_{\max}\), \(F_{\mathrm{wmax}}\), AUPR, IA-MAP, NDCG@5, and IA-R@5. All spokes report raw scores on a common scale, with higher values indicating better performance.}
\label{fig:dyna_reasoning}
\end{figure}

\section{Related Work}

\textbf{Sequence-based methods}.
Early approaches to protein function prediction relied on transferring annotations from homologous sequences with known functions, using tools such as BLAST~\citep{altschul1990basic}. Domain-based methods like InterProScan~\citep{jones2014interproscan} search for conserved sequence motifs associated with known functions. More recently, deep learning methods have replaced hand-crafted features with learned representations, including DeepGO~\citep{kulmanov2018deepgo} and DeepGOPlus~\citep{kulmanov2020deepgoplus}. ProLanGO~\citep{cao2017prolango} casts sequence-to-GO prediction as neural machine translation, while TALE~\citep{cao2021tale} jointly embeds protein sequences and hierarchical GO labels and DeepGO-SE~\citep{kulmanov2024semantic} incorporates ontology knowledge through approximate semantic entailment. ProtFun~\citep{talo2025protfun} combines protein language model embeddings with graph attention over a protein family network and InterPro signatures. GO-GPT~\citep{fallahpour2026bioreason} treats GO prediction as an autoregressive sequence generation task conditioned on protein language model embeddings, capturing hierarchical and cross-aspect dependencies among GO terms. Despite these advances, sequence-based predictors do not directly observe the protein's conformational ensemble at inference time.

\textbf{Structure-based methods}.
The availability of predicted structures from AlphaFold~\citep{jumper2021highly} and ESMFold~\citep{lin2023evolutionary} has enabled structure-based approaches at proteome scale. DeepFRI~\citep{gligorijevic2021structure} is a representative method that uses a graph convolutional network over protein contact maps, combined with sequence features from a language model, to predict GO terms. TransFun~\citep{boadu2023combining} combines transformer-based sequence embeddings with equivariant graph neural networks over AlphaFold-predicted structures, while FlexRibbon~\citep{zhu2026flexribbon} jointly pretrains sequence and structure representations and evaluates them on GO prediction. These methods have demonstrated that structural information can improve function prediction beyond what sequence alone provides. Earlier studies used molecular-dynamics ensembles and dynamics-derived fingerprints in function-prediction settings~\citep{glazer2009improving,hensen2012exploring}. SiamDiff~\citep{zhang2023siamdiff} learns protein representations by coupling diffusion trajectories from structurally correlated conformers, and DynamicsPLM builds learned protein representations from computational conformer ensembles~\citep{kalifa2026learning}. Deep Signature~\citep{qin2025deepsignature} characterizes molecular-dynamics trajectories through clustering of cooperative dynamics and signature transforms. Most structure-based predictors nevertheless operate on a single static structure per protein.

\textbf{LLM-based methods}.
A growing line of work uses large language models (LLMs) to generate natural language descriptions of protein functions, moving beyond fixed ontology vocabularies. Mol-Instructions~\citep{fang2024molinstructions} provides large-scale biomolecular instruction data that include protein-oriented prediction tasks. Prot2Text~\citep{abdine2024prot2text} combines protein sequences and structural graphs to generate functional descriptions, while its successor Prot2Text-V2~\citep{fei2025prot2text} generates such descriptions directly from sequence. ProtT3~\citep{liu2024prott3} couples a protein language model to a language model for protein captioning and question answering. STELLA~\citep{xiao2026stella} integrates sequence and single-structure representations with an LLM for functional annotation. ProteinChat~\citep{huo2024multi} adopts a multi-modal dialogue framework where users can interactively query a protein's function in natural language. InstructProtein~\citep{wang2024instructprotein} aligns protein and human language through knowledge-guided instruction tuning. OPUS-PLLM~\citep{lv2026opuspllm} instruction-tunes a sequence-conditioned generative model across GO prediction and other functional tasks. Recent work has also examined data leakage and biologically grounded evaluation in protein-to-text benchmarks~\citep{wu2025rethinking}, while PFUA frames protein-function understanding as tool-augmented, evidence-grounded reasoning~\citep{fan2026pfua}. ProtFunAgent~\citep{anonymous2025protfunagent} combines homology-guided retrieval with ontology-constrained decoding and an LLM cascade, while \emph{LLM Agent Based Protein Function Prediction}~\citep{zhapacamacho2025llmagent} refines GO predictions through tools and knowledge-augmented reasoning. BioReason-Pro~\citep{fallahpour2026bioreason} combines GO-GPT predictions with the protein sequence, organism information, and InterPro annotations in an LLM to generate structured functional rationales.

Deep Signature~\citep{qin2025deepsignature} models time-ordered MD trajectories and does not perform autoregressive GO prediction or GO-anchored language reasoning; DynamicsPLM~\citep{kalifa2026learning} learns conformational-ensemble representations for downstream prediction tasks, but not this GO-to-reasoning pipeline. GO-GPT and BioReason-Pro~\citep{fallahpour2026bioreason} provide autoregressive GO prediction and GO-anchored language reasoning, respectively, but do not condition GO generation on conformational ensembles. DynaReason-Pro's novelty is the combination of conformational-ensemble-conditioned autoregressive GO prediction and GO-anchored natural-language functional reasoning.

\section{Method}

DynaReason-Pro is a unified framework for ensemble-conditioned protein function prediction.
Its ontology-prediction module predicts a list of gene ontology (GO) terms for a protein represented by its sequence and conformational ensemble.
Its reasoning module generates functional descriptions and natural-language rationales grounded in the predicted GO terms, protein sequence, organism information, and InterPro annotations.

The ontology-prediction module of DynaReason-Pro predicts GO terms from a protein sequence and its conformational ensemble. Section~\ref{sec:method_problem} formulates aspect-specific GO prediction as conditional sequence generation, Section~\ref{sec:method_ensemble_encoder} converts the ensemble into residue-aligned structural context, and Section~\ref{sec:method_decoding} injects sequence-derived and ensemble-derived context tokens into the autoregressive GO decoder. The reasoning module, described in Section~\ref{sec:method_reasoning}, uses the predicted GO terms as ontology-level anchors for LLM-based functional reasoning, turning GO predictions into more explicit functional explanations.

\subsection{Ensemble-conditioned GO generation}
\label{sec:method_problem}

Given a protein sequence $x=(a_1,\ldots,a_L)$, let $C=\{c^{(1)},\ldots,c^{(K)}\}$ denote its unordered conformational ensemble, where each \(c^{(k)}\) represents one structural state. Here, conformational ensembles represent dynamics-related structural heterogeneity. GO annotations are organized into three aspects: Molecular Function (MF), Biological Process (BP), and Cellular Component (CC)~\citep{ashburner2000gene}. The function of a protein is represented as a sequence of GO-token sequence
\begin{equation}
\label{eq:go_sequence}
G^{(r)}=(g_1^{(r)},\ldots,g_T^{(r)}).
\end{equation}

The ensemble-conditioned GO prediction problem is formulated in Eq.~\eqref{eq:ensemble_go_factorization} as
\begin{equation}
\label{eq:ensemble_go_factorization}
p(G^{(r)}\mid x,C,r)
=
\prod_{t=1}^{T}
p(g_t^{(r)}\mid g_{<t}^{(r)},x,C,r).
\end{equation}

The goal is to construct a protein context that jointly represents the amino-acid sequence and the conformational ensemble. The key difficulty is that the ensemble is naturally indexed by structures, whereas autoregressive GO decoding requires a sequence of context tokens. We therefore convert the conformational ensemble into residue-aligned structural context before decoding.

\subsection{Residue-aligned ensemble-to-context translation}
\label{sec:method_ensemble_encoder}

We construct two residue-level views of the protein: a sequence view and an ensemble structural view. The sequence view is obtained by encoding the amino-acid sequence into residue-level representations,
\begin{equation}
\label{eq:sequence_encoder}
H_{\mathrm{seq}}
=
f_{\mathrm{seq}}(x)
\in
\mathbb{R}^{L\times d_{\mathrm{seq}}},
\end{equation}
followed by a projection into the decoder hidden space:
\begin{equation}
\label{eq:sequence_projection}
Z_{\mathrm{seq}}
=
P_{\mathrm{seq}}(H_{\mathrm{seq}})
\in
\mathbb{R}^{L\times d}.
\end{equation}

For the structural view, each conformation \(c^{(k)}\) is encoded into residue-level structural features:
\begin{equation}
\label{eq:structure_encoder}
H_{\mathrm{str}}^{(k)}
=
f_{\mathrm{str}}(x,c^{(k)})
=
[u_1^{(k)},\ldots,u_L^{(k)}]
\in
\mathbb{R}^{L\times d_{\mathrm{str}}},
\end{equation}
where
\begin{equation}
\label{eq:residue_feature}
u_\ell^{(k)}\in\mathbb{R}^{d_{\mathrm{str}}}
\end{equation}
denotes the structure-aware feature of residue \(\ell\) under conformation \(k\).

The conformational ensemble is initially structure-indexed: each \(H_{\mathrm{str}}^{(k)}\) describes all residues under one structural state. However, the decoder context is residue-indexed. We therefore re-index the ensemble by residue position. For each residue \(\ell\), we collect its features across all conformations:
\begin{equation}
\label{eq:residue_set}
U_\ell
=
\{u_\ell^{(1)},u_\ell^{(2)},\ldots,u_\ell^{(K)}\}.
\end{equation}
This gives a residue-indexed ensemble representation:
\begin{equation}
\label{eq:residue_indexed_ensemble}
U=(U_1,\ldots,U_L),
\qquad
U\in\mathbb{R}^{K\times L\times d_{\mathrm{str}}}.
\end{equation}
Each \(U_\ell\) is a set of alternative structural states for the same residue. This representation preserves the alignment between the conformational ensemble and the original amino-acid sequence, while avoiding a global protein-level compression of the ensemble.

After re-indexing, each \(U_\ell\) is still a set rather than a decoder token. We therefore learn a residue-wise set-to-token translator \(E_\psi\) that maps each conformational set into one structural context token:
\begin{equation}
\label{eq:set_to_token}
z_{\mathrm{str},\ell}
=
E_\psi(U_\ell)
\in
\mathbb{R}^{d}.
\end{equation}

The conformation-specific residue features are first projected into the decoder hidden space:
\begin{equation}
\label{eq:input_projection}
X_{\ell,0}
=
P_{\mathrm{in}}(U_\ell)
\in
\mathbb{R}^{K\times d}.
\end{equation}
For each residue \(\ell\), we apply self-attention over its \(K\) conformer tokens:
\begin{equation}
\label{eq:self_attention}
\bar X_{\ell}
=
\mathrm{SelfAttn}_{\psi}(X_{\ell,0},X_{\ell,0},X_{\ell,0})
\in
\mathbb{R}^{K\times d}.
\end{equation}
Here \(\mathrm{SelfAttn}_{\psi}\) denotes multi-head self-attention over the \(K\) conformer tokens of residue \(\ell\), implemented with standard pre-layer-normalized Transformer blocks~\citep{vaswani2017attention,lee2019set,xiong2020layer}.

We then compress this encoded conformer set into a single residue-level structural token by learned-query attention pooling, with attention weights and pooled output defined in Eqs.~\eqref{eq:pool_weights} and~\eqref{eq:pool_output}:
\begin{equation}
\label{eq:pool_weights}
\alpha_{\ell}
=
\mathrm{softmax}
\left(
\frac{
(q_{\mathrm{pool}}W_Q)
(\bar X_{\ell}W_K)^\top
}{
\sqrt{d_h}
}
\right)
\in
\mathbb{R}^{1\times K},
\end{equation}
\begin{equation}
\label{eq:pool_output}
z_{\mathrm{str},\ell}
=
\mathrm{LN}_{\mathrm{out}}
\left(
\alpha_{\ell}\bar X_{\ell}W_V
\right)
\in
\mathbb{R}^{d},
\end{equation}
where \(q_{\mathrm{pool}}\in\mathbb{R}^{1\times d}\) is a learnable pooling query and \(d_h\) is the attention-head dimension. In practice, pooling is implemented as multi-head attention; the equations above show the single-head form for clarity. Applying the same translator to all residues gives
\begin{equation}
\label{eq:structural_context}
Z_{\mathrm{str}}
=
(z_{\mathrm{str},1},\ldots,z_{\mathrm{str},L})
\in
\mathbb{R}^{L\times d}.
\end{equation}

Thus, the structural branch produces a sequence of structural context tokens aligned with the sequence tokens, rather than a single global structural embedding. The final fusion is therefore performed after the ensemble has been converted into residue-aligned context.

\subsection{Dual-view autoregressive decoding}
\label{sec:method_decoding}

The decoder receives a dual-view context composed of sequence-derived and ensemble-derived residue tokens. Before concatenation, we add learned view-specific type embeddings to distinguish the two context sources:
\begin{equation}
\label{eq:type_embeddings}
\tilde Z_{\mathrm{seq}}
=
Z_{\mathrm{seq}}+e_{\mathrm{seq}},
\qquad
\tilde Z_{\mathrm{str}}
=
Z_{\mathrm{str}}+e_{\mathrm{str}},
\end{equation}
\begin{equation}
\label{eq:dual_view_context}
Z
=
[
\tilde Z_{\mathrm{seq}};
\tilde Z_{\mathrm{str}}
]
\in
\mathbb{R}^{2L\times d}.
\end{equation}
Here \(e_{\mathrm{seq}},e_{\mathrm{str}}\in\mathbb{R}^{d}\) are learned type embeddings for sequence-derived and ensemble-derived structural context tokens.

The dual-view context \(Z\) in Eq.~\eqref{eq:dual_view_context} is injected as prefix context tokens in the autoregressive decoder, not through a separate encoder--decoder cross-attention module. Within this prefix, sequence-derived tokens are followed by ensemble-derived structural tokens, and the complete context prefix is placed before the GO-token sequence. Therefore, each GO decoding position can attend to the sequence and structural context as well as previous GO tokens, while causal masking is preserved among GO tokens.

GO generation is parameterized as
\begin{equation}
\label{eq:decoder_conditioning}
p(G^{(r)}\mid x,C,r)
=
p_{\mathrm{dec}}
\left(
G^{(r)}
\mid
Z,r
\right).
\end{equation}
At decoding step \(t\), the token distribution is
\begin{equation}
\label{eq:token_distribution}
p(g_t^{(r)}\mid g_{<t}^{(r)},Z,r)
=
\mathrm{softmax}
\left(
W_{\mathrm{go}}h_t^{\mathrm{ens}}
\right),
\end{equation}
where \(h_t^{\mathrm{ens}}\) is the decoder state under the dual-view prefix context.

Within each aspect, GO targets are ordered deterministically by ascending GO identifier and enclosed by aspect-specific start and end tokens. Training uses teacher forcing on aspect-specific GO-token sequences. The objective is the token-level cross entropy in Eq.~\eqref{eq:training_loss}, with padding tokens ignored:
\begin{equation}
\label{eq:training_loss}
\mathcal{L}
=
-
\frac{1}{N_{\mathrm{tok}}}
\sum_i
\sum_t
\mathbf{1}
[
g_{i,t}^{(r_i)}\neq \mathrm{PAD}
]
\log
p
\left(
g_{i,t}^{(r_i)}
\mid
g_{i,<t}^{(r_i)},
Z_i,
r_i
\right),
\end{equation}
where \(i\) indexes protein-aspect examples, \(r_i\in\{\mathrm{MF},\mathrm{BP},\mathrm{CC}\}\), and \(N_{\mathrm{tok}}\) is the number of non-padding target tokens.

At inference time, the model constructs the dual-view context \(Z\) from the input sequence and its conformational ensemble, and generates an aspect-specific GO-token sequence \(\hat{G}^{(r)}\) for each aspect \(r\).

\subsection{Ontology-anchored functional reasoning}
\label{sec:method_reasoning}

The reasoning module of DynaReason-Pro turns predicted GO terms into functional reasoning. We reuse the BioReason-Pro reasoning pipeline~\citep{fallahpour2026bioreason}; DynaReason-Pro changes the functional anchors supplied to it by predicting GO terms from conformational ensembles. Given the GO terms generated by the ontology-prediction module, an LLM-based reasoning module uses them as ontology-level anchors together with protein-level biological context to generate a natural-language functional explanation. The GO terms provide compact functional evidence, while the reasoning module organizes this evidence into a more interpretable account of molecular function, biological process, and cellular context.

This design separates GO grounding from explanation generation: the ontology-prediction module of DynaReason-Pro provides ensemble-conditioned GO anchors from conformational ensembles, and the reasoning module turns these anchors into coherent functional descriptions.

\section{Experiment}
\label{sec:experiments}

We evaluate DynaReason-Pro along two complementary axes. First, we benchmark its ensemble-conditioned ontology-prediction module on the controlled PED ensemble test split, using the same propagated non-root GO evaluation protocol for all methods. Second, we evaluate whether these ensemble-aware GO predictions provide better ontology-level anchors for LLM-based functional reasoning. We further analyze the learned model through visualization and ablation studies to examine whether the gains come from meaningful ensemble information. The evaluation also includes conformational-variability stratification and case studies. Detailed implementation and reasoning-module settings are provided in Appendix~\ref{app:implementation}.

\subsection{Evaluation setup}
\label{sec:setup}

\textbf{Benchmark construction}. We construct an IDP ensemble GO-prediction benchmark from the Protein Ensemble Database (PED)~\citep{ghafouri2024ped}. PED is a curated resource for structural ensembles of intrinsically disordered proteins and disordered regions. Each retained sample corresponds to one PED ensemble entry, augmented with UniProt-mapped Gene Ontology (GO) annotations and a matched AlphaFold DB (AFDB)~\citep{varadi2022alphafold} single-structure model for single-structure baselines and split building.

For simplicity, we use a fixed ensemble size of \(K=32\) for all retained entries in the PED database, which balances data set coverage and computational efficiency. For entries with more than 32 valid conformers, we sample 32 conformers randomly. Entries with fewer than 32 valid conformers are excluded.\footnote{After filtering, 5,902 out of 6,239 usable PED candidates have at least 32 unique conformers, preserving approximately 94.6\% of candidate entries.} This fixed-size design avoids padding variable-size ensembles while keeping residue-wise set-attention over the conformer dimension simple. We extract the ESM2 sequence representation and ESM3 structure-conditioned residue embeddings for the 32 PED conformers and for the matched AFDB single structure~\citep{lin2023evolutionary,hayes2025simulating}. GO annotations are organized by the three standard GO categories: Molecular Function (MF), Biological Process (BP), and Cellular Component (CC)~\citep{ashburner2000gene}.

To avoid leakage and overly optimistic evaluation, we use a strict split based on both sequence and structure. We first cluster protein sequences with MMseqs2 at 50\% sequence identity~\citep{steinegger2017mmseqs2} and cluster AFDB single structures using Foldseek~\citep{vankempen2024foldseek}. We then construct joint sequence-structure components so that proteins connected by either sequence similarity or structural similarity are assigned to the same split. Validation and test proteins are additionally required to be disjoint from the pretraining data used for the sequence-only ontology predictor at the exact protein, sequence-cluster, and AFDB structure-cluster levels. Our final dataset contains 5,902 PED entry/ensemble records: 5,654 for training, 130 for validation, and 118 for testing. Data provenance, accession and sequence counts, and annotation policy are provided in Appendix~\ref{app:provenance}. GO target construction is detailed in Appendix~\ref{app:targets}.

\textbf{Evaluation metrics}.
All metrics are computed on the same test set and the same propagated non-root GO-term universe. Missing prediction scores are filled with zero. Our metrics include threshold-swept, precision-recall, and semantic-distance metrics commonly used in CAFA-style GO function prediction evaluation~\citep{radivojac2013large,clark2013information,kulmanov2024semantic}. This includes \(F_{\max}\), \(F_{\mathrm{wmax}}\) (weighted \(F_{\max}\)), aspect-specific \(F_{\max}/F_{\mathrm{wmax}}\), high-information-content \(F_{\max}/F_{\mathrm{wmax}}\), AUPR, and \(S_{\min}\). We also report information-aware and top-$k$ retrieval criteria, including IA-MAP, NDCG, Jaccard, and IA-R. Higher values are better for all metrics except \(S_{\min}\). Detailed definitions are provided in Appendix~\ref{app:metrics}.

\textbf{Baselines}. We compare our ensemble-conditioned model with the internal sequence-only ontology predictor, external learned predictors, and simple statistical controls. The external predictors are DeepGO, the sequence-based DeepFRI-CNN, the AFDB-single-structure-based DeepFRI-GCN, and the LLM-based BioReason-Pro. The statistical controls reported in the main tables are an aspect-frequency GO prior, length-based 5NN label transfer, and amino-acid-composition 5NN label transfer. Baseline details are provided in Appendix~\ref{app:baselines}.
The ablation study compares sequence-only prediction, benchmark-trained AFDB single-structure conditioning, independent 32-conformer averaging, and fixed-split ESM2 sequence and ESM3 AFDB-structure retrieval controls; details are provided in Appendices~\ref{app:baselines} and~\ref{app:ablation}.

\subsection{Results}
\label{sec:results}

\subsubsection{Comparison with baselines}
\label{sec:benchmark}

Table~\ref{tab:cafa_metrics} compares DynaReason-Pro with learned sequence and structure predictors and with statistical controls using CAFA-style metrics, while Table~\ref{tab:ranking_metrics} provides other statistical and top-$k$ retrieval evaluations. Compared with the learned baselines in Table~\ref{tab:cafa_metrics}, DynaReason-Pro attains the highest overall \(F_{\max}\) (0.6045 versus 0.439 for DeepFRI-GCN) and \(F_{\mathrm{wmax}}\) (0.5172 versus 0.357 for DeepFRI-GCN), and records the highest value on all 16 information-aware and top-$k$ metrics in Table~\ref{tab:ranking_metrics}. Performance remains aspect dependent, with Biological Process presenting the most difficult branch. Prediction coverage and multi-seed results are reported in Appendices~\ref{app:bp} and~\ref{app:seeds}.

\begin{table}[t]
\centering
\caption{CAFA-style metrics on the PED test set. Arrows indicate the preferred metric direction. \textbf{Bold} indicates the best method for each metric.}
\label{tab:cafa_metrics}
\setlength{\tabcolsep}{0.9pt}
\begin{tabular}{@{}llccccccccc@{}}
\toprule
\multirow{2}{*}{Metric} & \multirow{2}{*}{Aspect}
& \multicolumn{3}{c}{Statistics}
& \multicolumn{3}{c}{Sequence}
& & & \\
\cmidrule(lr){3-5}\cmidrule(lr){6-8}
&
& \rotatebox{60}{GO prior}
& \rotatebox{60}{length 5NN}
& \rotatebox{60}{AA comp. 5NN}
& \rotatebox{60}{\shortstack[l]{Sequence-only\\predictor}}
& \rotatebox{60}{DeepGO}
& \rotatebox{60}{DeepFRI-CNN}
& \rotatebox{60}{\shortstack[l]{Structure\\DeepFRI-GCN}}
& \rotatebox{60}{\shortstack[l]{LLM-based\\BioReason-Pro}}
& \rotatebox{60}{\shortstack[l]{Ensemble-aware\\DynaReason-Pro}} \\
\midrule
\multirow{5}{*}{\(F_{\max}\uparrow\)}
& Overall & 0.4066 & 0.3392 & 0.4247 & 0.4226 & 0.4261 & 0.3983 & 0.4390 & 0.4251 & \textbf{0.6045} \\
& MF & 0.2944 & 0.2301 & 0.3050 & 0.3306 & 0.4470 & 0.4104 & 0.4081 & 0.3223 & \textbf{0.4814} \\
& BP & 0.2104 & 0.0852 & 0.1885 & 0.3020 & 0.3071 & 0.2948 & 0.2214 & 0.3064 & \textbf{0.3163} \\
& CC & 0.4827 & 0.4300 & 0.5120 & 0.4634 & 0.4537 & 0.4663 & 0.5424 & 0.4579 & \textbf{0.7004} \\
& High-IA & 0.3617 & 0.1991 & 0.3194 & 0.2337 & 0.3481 & 0.3318 & 0.3430 & 0.2356 & \textbf{0.4897} \\
\midrule
\multirow{5}{*}{\(F_{\mathrm{wmax}}\uparrow\)}
& Overall & 0.3370 & 0.2244 & 0.3140 & 0.2830 & 0.3318 & 0.3235 & 0.3566 & 0.2890 & \textbf{0.5172} \\
& MF & 0.2499 & 0.1636 & 0.2847 & 0.1902 & 0.4090 & 0.3398 & 0.3373 & 0.1704 & \textbf{0.4424} \\
& BP & 0.2469 & 0.0675 & 0.1852 & 0.2877 & 0.2636 & 0.2684 & 0.1903 & \textbf{0.3014} & 0.2745 \\
& CC & 0.3895 & 0.2731 & 0.3672 & 0.2756 & 0.3166 & 0.3645 & 0.4500 & 0.2812 & \textbf{0.5924} \\
& High-IA & 0.3492 & 0.1877 & 0.3092 & 0.2282 & 0.3344 & 0.3093 & 0.3256 & 0.2283 & \textbf{0.4834} \\
\midrule
\multicolumn{2}{l}{AUPR\(\uparrow\)} & 0.1896 & 0.1318 & 0.1963 & 0.1949 & 0.3475 & 0.3049 & 0.3317 & 0.0903 & \textbf{0.4012} \\
\multicolumn{2}{l}{\(S_{\min}\downarrow\)} & 18.3791 & 25.1900 & 18.0098 & 23.5443 & 17.8365 & 18.7732 & 18.4594 & 27.4941 & \textbf{16.6653} \\
\bottomrule
\end{tabular}
\end{table}

\begin{table}[t]
\centering
\caption{Information-aware and top-\(k\) retrieval metrics on the PED test set for the same methods in Table~\ref{tab:cafa_metrics}. All metrics are reported with \(\uparrow\), indicating that higher is better. \textbf{Bold} indicates the best method for each metric. N@\(k\) = NDCG@\(k\), J@\(k\) = Jaccard@\(k\), and R@\(k\) = IA-R@\(k\).}
\label{tab:ranking_metrics}
\setlength{\tabcolsep}{1.5pt}
\begin{tabular}{@{}lccccccccc@{}}
\toprule
\multirow{2}{*}{Metric}
& \multicolumn{3}{c}{Statistics}
& \multicolumn{3}{c}{Sequence}
& & & \\
\cmidrule(lr){2-4}\cmidrule(lr){5-7}
& \rotatebox{60}{GO prior}
& \rotatebox{60}{length 5NN}
& \rotatebox{60}{AA comp. 5NN}
& \rotatebox{60}{\shortstack[l]{Sequence-only\\predictor}}
& \rotatebox{60}{DeepGO}
& \rotatebox{60}{DeepFRI-CNN}
& \rotatebox{60}{\shortstack[l]{Structure\\DeepFRI-GCN}}
& \rotatebox{60}{\shortstack[l]{LLM-based\\BioReason-Pro}}
& \rotatebox{60}{\shortstack[l]{Ensemble-aware\\DynaReason-Pro}} \\
\midrule
IA-MAP\(\uparrow\) & 0.3494 & 0.2063 & 0.3412 & 0.2068 & 0.3294 & 0.3780 & 0.3951 & 0.2110 & \textbf{0.4309} \\
N@1\(\uparrow\) & 0.0845 & 0.0739 & 0.0796 & 0.0729 & 0.1081 & 0.1395 & 0.1553 & 0.1002 & \textbf{0.2940} \\
N@3\(\uparrow\) & 0.1029 & 0.1349 & 0.1777 & 0.1447 & 0.1260 & 0.1972 & 0.1988 & 0.1510 & \textbf{0.3557} \\
N@5\(\uparrow\) & 0.1096 & 0.1618 & 0.2026 & 0.1515 & 0.1624 & 0.2633 & 0.2815 & 0.1646 & \textbf{0.3704} \\
N@10\(\uparrow\) & 0.2072 & 0.1863 & 0.2388 & 0.1626 & 0.2055 & 0.3068 & 0.3208 & 0.1751 & \textbf{0.3779} \\
J@3\(\uparrow\) & 0.0862 & 0.0959 & 0.1213 & 0.1500 & 0.1734 & 0.1492 & 0.1797 & 0.1418 & \textbf{0.3231} \\
J@5\(\uparrow\) & 0.1254 & 0.1218 & 0.1752 & 0.1947 & 0.2185 & 0.2270 & 0.2542 & 0.1771 & \textbf{0.3757} \\
J@10\(\uparrow\) & 0.2292 & 0.1292 & 0.2034 & 0.2415 & 0.2219 & 0.2468 & 0.2452 & 0.2396 & \textbf{0.4342} \\
R@1\(\uparrow\) & 0.0363 & 0.0522 & 0.0412 & 0.0361 & 0.0552 & 0.0712 & 0.0903 & 0.0637 & \textbf{0.2235} \\
R@3\(\uparrow\) & 0.0900 & 0.1562 & 0.2086 & 0.1545 & 0.1127 & 0.1984 & 0.1956 & 0.1468 & \textbf{0.3554} \\
R@5\(\uparrow\) & 0.1122 & 0.2226 & 0.2670 & 0.1793 & 0.1971 & 0.3527 & 0.3884 & 0.1881 & \textbf{0.3945} \\
CC IA-MAP\(\uparrow\) & 0.4280 & 0.2875 & 0.4309 & 0.2331 & 0.3452 & 0.4357 & 0.4758 & 0.2343 & \textbf{0.5479} \\
CC N@1\(\uparrow\) & 0.1116 & 0.1106 & 0.1114 & 0.0656 & 0.1149 & 0.1618 & 0.2094 & 0.1213 & \textbf{0.4142} \\
CC N@3\(\uparrow\) & 0.1383 & 0.1974 & 0.2516 & 0.1707 & 0.1278 & 0.2345 & 0.2567 & 0.1913 & \textbf{0.4739} \\
CC N@5\(\uparrow\) & 0.1343 & 0.2357 & 0.2789 & 0.1778 & 0.1677 & 0.3102 & 0.3706 & 0.2029 & \textbf{0.4869} \\
CC N@10\(\uparrow\) & 0.2692 & 0.2672 & 0.3213 & 0.1868 & 0.2159 & 0.3606 & 0.4028 & 0.2101 & \textbf{0.4924} \\
\bottomrule
\end{tabular}
\end{table}

\subsubsection{Ablation study}
\label{sec:ablation_results}

\paragraph{Overall ablation.}
We report \(F_{\max}\) and \(F_{\mathrm{wmax}}\), the two primary aggregate metrics in our evaluation. We perform an ablation study to test the contribution of structural information and the way conformational ensembles are used. The sequence-only baseline uses only the original protein sequence representation. The AFDB single-structure variant adds one AlphaFold DB structure per protein as a residue-level structural context, testing whether a conventional single folded structure improves GO prediction. The PED 32 avg-single baseline evaluates the same single-structure model independently on 32 PED conformers and averages the resulting GO scores, testing whether naive ensemble averaging is sufficient. Our final model instead encodes all 32 conformers jointly through a residue-aligned ensemble encoder, producing an ensemble-aware structural context for GO generation.

\begin{figure*}[t]
\centering
\includegraphics[width=\textwidth]{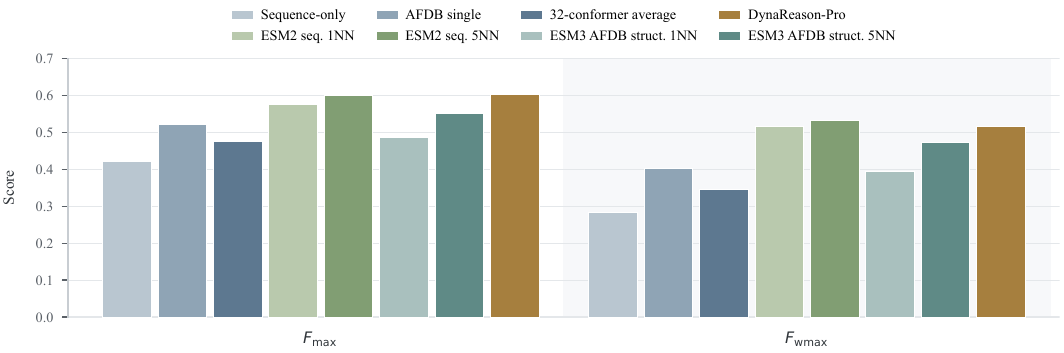}
\caption{\textbf{Overall ablation across architectural and retrieval variants.} \(F_{\max}\) and \(F_{\mathrm{wmax}}\) compare sequence-only prediction, AFDB single-structure conditioning, naive averaging over 32 PED conformers, ESM2 sequence retrieval, ESM3 AFDB-structure retrieval, and DynaReason-Pro.}
\label{fig:ablation}
\end{figure*}

Figure~\ref{fig:ablation} shows that DynaReason-Pro improves both \(F_{\max}\) and \(F_{\mathrm{wmax}}\) over sequence-only prediction, AFDB single-structure conditioning, and naive averaging over 32 PED conformers, demonstrating the benefit of learned ensemble integration. ESM-based retrieval remains competitive on the aggregate test set: ESM2 sequence 5NN approaches DynaReason-Pro in \(F_{\max}\) and attains a slightly higher \(F_{\mathrm{wmax}}\). We therefore stratify performance by conformational variability to examine the benefit of ensemble modeling for proteins with greater dynamics-related structural heterogeneity. Additional ensemble controls and conformer-count, sampling, and order analyses are reported in Appendices~\ref{app:controlled}--\ref{app:negative}.

\paragraph{Variability-dependent ablation.}
\label{sec:variability_results}

We stratify the 118 test records by mean pairwise C$\alpha$ TM-score across 32 conformers, producing four approximately equal quartiles (30/30/29/29 records) from Q1 (lowest variability) to Q4 (highest variability)~\citep{zhang2004scoring}. Lower mean TM-score denotes weaker within-ensemble structural agreement. This analysis therefore tests whether the contribution of ensemble conditioning increases with conformational heterogeneity.

\begin{figure}[t]
\centering
\includegraphics[width=0.96\textwidth,trim=7bp 8.5bp 7bp 9bp,clip]{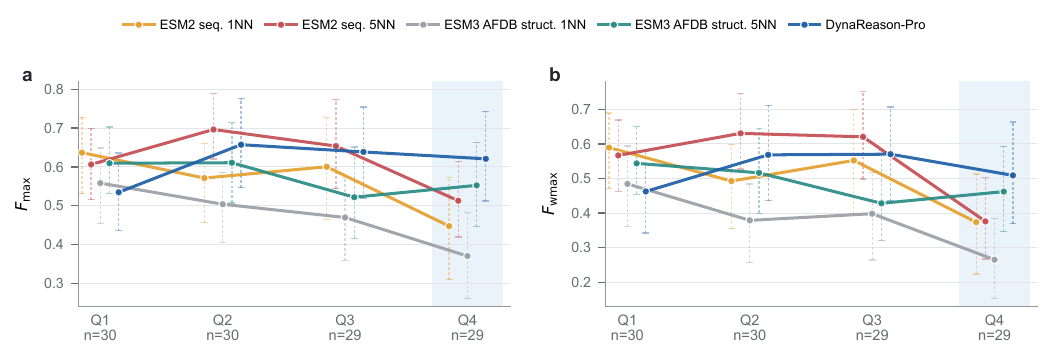}
\caption{\textbf{DynaReason-Pro has its largest relative advantage in the highest-variability quartile.} \textbf{a}, \(F_{\max}\); \textbf{b}, \(F_{\mathrm{wmax}}\). Both metrics are shown for DynaReason-Pro and four fixed-split retrieval variants across TM-score quartiles (Q1--Q4; $n=30,30,29,29$). Dashed error bars show 95\% bootstrap confidence intervals from 10,000 within-quartile resamples (2.5th--97.5th percentiles); markers are horizontally offset for visibility. DynaReason-Pro leads all four variants in Q4.}
\label{fig:variability_esm2_trend}
\end{figure}

Figure~\ref{fig:variability_esm2_trend} shows that DynaReason-Pro leads all four retrieval variants in Q4, reaching 0.6209 \(F_{\max}\) and 0.5091 \(F_{\mathrm{wmax}}\). Relative to ESM2 sequence 5NN, its Q1--Q4 margins are $-0.0718$, $-0.0392$, $-0.0151$, and $+0.1080$ for \(F_{\max}\), and $-0.1041$, $-0.0620$, $-0.0501$, and $+0.1329$ for \(F_{\mathrm{wmax}}\). Relative to ESM2 sequence 5NN, the observed margins increase monotonically across the four conformational-variability quartiles, with the largest gains in Q4. Covariate-adjusted, continuous-variability, and matching analyses are reported in Appendix~\ref{app:variability}.

\subsubsection{Visualization}
\label{sec:visualization}

\begin{figure}[!htbp]
\centering
\includegraphics[width=\textwidth,trim=55.5bp 434.5bp 56bp 54bp,clip]{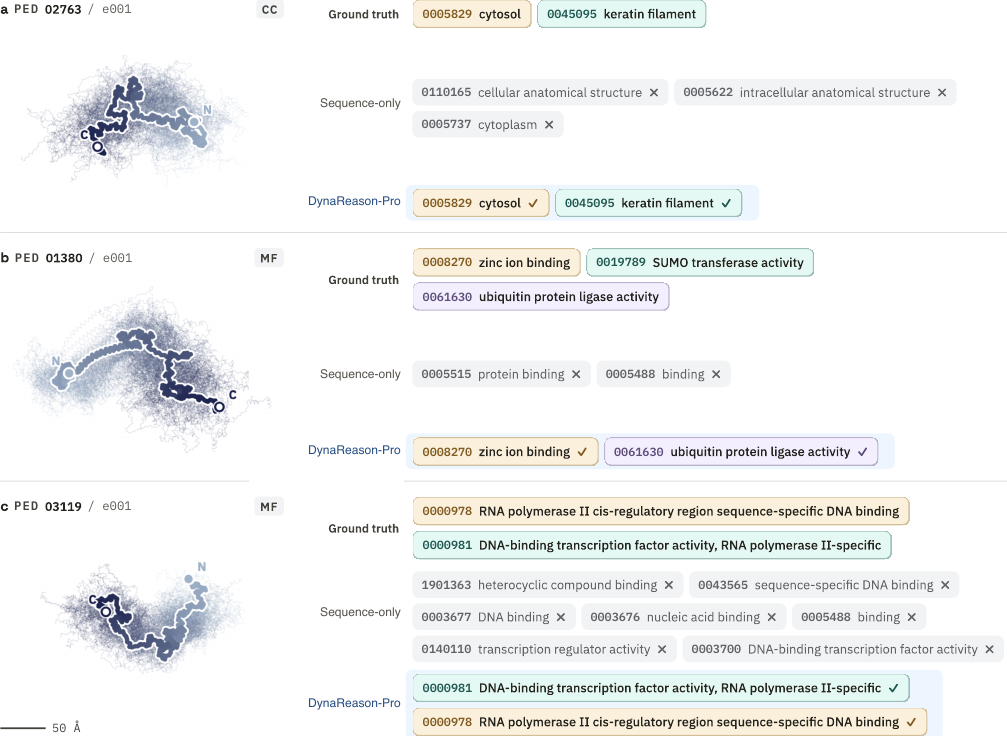}
\caption{\textbf{Ensemble conditioning removes incorrect predictions and recovers specific GO functions.} Across three PED proteins, DynaReason-Pro replaces broad sequence-only outputs with exact reference terms: (a) cytosol and keratin filament, (b) zinc ion binding and ubiquitin protein ligase activity, and (c) two RNA polymerase II-specific DNA-binding functions.}
\label{fig:raw_go_audit}
\end{figure}

\paragraph{Term-level case studies.}
Figure~\ref{fig:raw_go_audit} shows a direct shift from broad, incorrect predictions to specific ground-truth functions. For PED02763\_e001, the sequence-only model emits three unmatched cellular terms, whereas DynaReason-Pro removes these errors and recovers the exact cytosol and keratin filament annotations. For PED01380\_e001, it replaces generic protein-binding predictions with zinc ion binding and ubiquitin protein ligase activity. The contrast is strongest for PED03119\_e001: seven broad sequence-only predictions spanning generic binding and transcriptional regulation are eliminated, while both RNA polymerase II-specific DNA-binding terms are recovered exactly. Across all three proteins, conformational-ensemble conditioning simultaneously suppresses false predictions and increases recovery of specific reference terms.

\begin{figure*}[t]
\centering
\includegraphics[width=\textwidth,trim=13.5bp 22.5bp 31bp 3.5bp,clip]{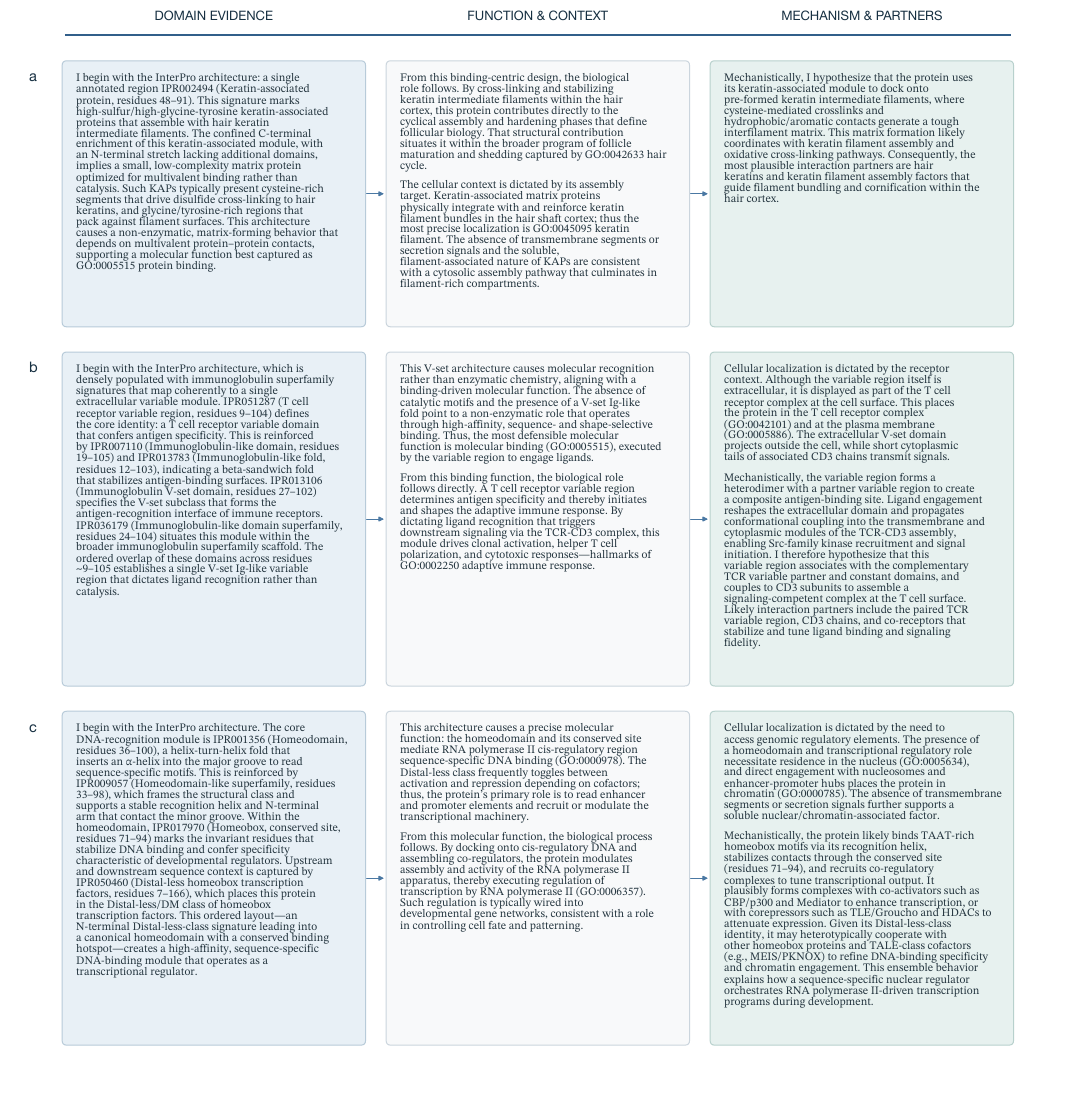}
\caption{\textbf{DynaReason-Pro resolves domain architecture into candidate protein-specific mechanisms.} Panel (a) shows the complete trace for PED02763\_e001, panel (b) for PED01773\_e001, and panel (c) for PED03119\_e001. Within each panel, the full trace is reflowed into three semantic blocks: domain evidence; function and cellular context; and mechanisms and candidate interaction partners.}
\label{fig:trace_audit_full}
\end{figure*}

\paragraph{Reasoning traces.}
Figure~\ref{fig:trace_audit_full} demonstrates protein-specific reasoning across three distinct domain architectures. For PED02763\_e001, DynaReason-Pro connects the cysteine-rich keratin-associated module to multivalent keratin binding, hair-cycle function, keratin-filament localization, and hair-keratin interaction partners. For PED01773\_e001, overlapping immunoglobulin V-set signatures are resolved into antigen recognition, adaptive immune response, localization to the T-cell receptor complex and plasma membrane, and coupling to TCR/CD3 partners. For PED03119\_e001, the conserved Distal-less homeodomain is traced to RNA polymerase II cis-regulatory DNA binding, nuclear and chromatin localization, developmental transcriptional control, and recruitment of transcriptional cofactors. These traces show that DynaReason-Pro can turn domain architecture and ontology anchors into candidate molecular mechanisms, cellular roles, and testable interaction hypotheses rather than isolated functional labels. Quantitative evaluation of functional reasoning is reported in Appendix~\ref{app:reasoning}.

\section{Discussion and Conclusion}
\label{sec:discussion}

DynaReason-Pro links ensemble-aware ontology prediction with ontology-grounded functional reasoning. The term-level and trace visualizations complement these aggregate results by showing how some improvements correspond to more specific raw GO emissions and by making the evidence-to-output trace inspectable.

DynaReason-Pro's clearest gains occur in the high-conformational-variability regime.

More broadly, DynaReason-Pro establishes a route from dynamics-related structural heterogeneity encoded in conformational ensembles to ontology prediction and onward to natural-language functional reasoning. Extending this route across broader protein classes and complementary sources of conformational ensembles may move protein function prediction beyond static annotation toward explanations grounded in changing molecular states.


\clearpage
\bibliography{refs}
\bibliographystyle{tmlr}




    \appendix

\section{Benchmark Provenance, Duplicate Handling, and Annotation Policy}
\label{app:provenance}

The benchmark unit is a PED entry/ensemble record. The benchmark contains 5,902 records, corresponding to 5,760 distinct UniProt accessions and 5,744 distinct model-input sequences. The final split contains 5,654 training, 130 validation, and 118 test records.

PED-to-UniProt mappings are taken directly from PED construct-fragment accession metadata. We do not introduce a new sequence-similarity search for these mappings or remap isoforms to canonical accessions. No accession group or exact model-input-sequence group spans training, validation, and test. In addition, the sequence--structure component split described in the main text separates records linked through MMseqs2 sequence clustering or Foldseek structural clustering.

GO cross-references were acquired on 8--9 April 2026 UTC and parsed without an evidence-code whitelist or blacklist. Consequently, electronically inferred annotations (IEA) were not excluded. This annotation policy is shared by the benchmark-controlled comparisons.

\begin{table}[htbp]
\centering
\small
\caption{Retained benchmark records and distinct identities. Counts of accessions and sequences refer to distinct values.}
\label{tab:provenance_counts}
\begin{tabular}{lr}
\toprule
Quantity & Count \\
\midrule
PED entry/ensemble records & 5,902 \\
Distinct UniProt accessions & 5,760 \\
Distinct model-input sequences & 5,744 \\
Training / validation / test records & 5,654 / 130 / 118 \\
Accession groups spanning splits & 0 \\
Exact model-input-sequence groups spanning splits & 0 \\
\bottomrule
\end{tabular}
\end{table}

\section{GO Target Construction}
\label{app:targets}

Within each aspect, targets are constructed from the existing GO-identifier lists and ordered deterministically by ascending GO identifier. Training targets are not expanded with ancestor terms.

Each aspect has its own start and end token. Teacher forcing uses the appropriate aspect-start token and shifted GO-token targets ending with that aspect's end token. Ancestor propagation is applied only during evaluation, together with aspect-root exclusion (Appendix~\ref{app:metrics}).

\section{Controlled Baseline Implementation and Aggregation Comparisons}
\label{app:controlled}
\label{app:ablation}

\paragraph{Sequence-only predictor.}
This control uses the pretrained sequence representation without structural input and provides the reference for measuring the contribution of structural conditioning.

\paragraph{AFDB single.}
    This variant uses the amino-acid sequence together with one matched AlphaFold DB single-structure model~\citep{varadi2022alphafold}. Per-residue ESM3 structural embeddings from the AFDB structure are used as a structural prefix. This tests whether a conventional single-structure representation improves GO prediction. The AFDB-single control is benchmark-trained on the same PED training/validation split.
    
    \paragraph{Independent 32-conformer average.}
This variant applies the single-structure predictor independently to each of the 32 PED conformers and averages the resulting GO scores. It tests whether multiple structures are sufficient without learned cross-conformer aggregation.

\paragraph{DynaReason-Pro ensemble conditioning.}
The full model jointly encodes 32 PED conformers with the residue-aligned ensemble encoder and fuses the resulting structural context with sequence context for autoregressive GO generation. This isolates the contribution of learned aggregation over conformational states.

DeepSets and attention-pooling controls test simpler learned aggregation alternatives on the benchmark (Table~\ref{tab:controlled_ensemble}). The full encoder exceeds DeepSets by \(0.01992/0.01795\) and attention pooling by \(0.04664/0.05774\) in \(F_{\max}/F_{\mathrm{wmax}}\). These comparisons favor learned cross-conformer integration over the tested simpler pooling alternatives. The ensemble perturbation and robustness analyses use a common reference model, with \(F_{\max}/F_{\mathrm{wmax}}=0.62267/0.52527\). In the main benchmark comparison, the full ensemble attains \(0.6045/0.5172\), AFDB single attains \(0.5209/0.4019\), and independent 32-conformer averaging attains \(0.4759/0.3471\).

\begin{table}[t]
\centering
\caption{\textbf{Controlled evidence for protein-specific ensemble conditioning.} The full, mismatched, and repeat-single inputs are evaluated with the same ensemble reference model. DeepSets and attention pooling are trained ensemble alternatives. Repeat-single ranges summarize five choices of one conformer duplicated 32 times.}
\label{tab:controlled_ensemble}
\begin{tabular}{@{}lcc@{}}
\toprule
Condition or aggregation & $F_{\max}$ & $F_{\mathrm{wmax}}$\\
\midrule
Full set-attention ensemble & \textbf{0.62267} & \textbf{0.52527}\\
Mismatched ensemble & 0.42431 & 0.25065\\
Repeated single conformer (range) & 0.50902--0.55238 & 0.38598--0.44531\\
DeepSets & 0.60275 & 0.50732\\
Attention pooling & 0.57603 & 0.46753\\
\bottomrule
\end{tabular}
\end{table}

\section{Conformer-Count, Subset, Sampling, and Order Robustness}
\label{app:robustness}

\paragraph{Conformer-count sensitivity.}
We compare trained \(K=8\) and \(K=16\) models with the \(K=32\) reference. Table~\ref{tab:conformer_count} summarizes sensitivity to the training conformer count.

\begin{table}[htbp]
\centering
\small
\caption{Conformer-count sensitivity for trained \(K=8\), \(K=16\), and \(K=32\) models.}
\label{tab:conformer_count}
\begin{tabular}{lrr}
\toprule
Conformer count & \(F_{\max}\) & \(F_{\mathrm{wmax}}\) \\
\midrule
\(K=8\) & 0.571077 & 0.454456 \\
\(K=16\) & 0.585052 & 0.458119 \\
\(K=32\) reference & 0.622673 & 0.525268 \\
\bottomrule
\end{tabular}
\end{table}

\paragraph{Random subsets.}
Five random subsets of the 32-conformer ensemble were evaluated at inference with the fixed ensemble reference model for each of \(K=8\) and \(K=16\). The mean and standard deviation across subsets are shown in Table~\ref{tab:random_subsets}. Variation is smaller for \(K=16\), supporting robustness to the identities of selected conformers.

\begin{table}[htbp]
\centering
\small
\caption{Random-subset robustness. Values are means \(\pm\) standard deviations over five subsets per conformer count.}
\label{tab:random_subsets}
\begin{tabular}{lrr}
\toprule
Subset size & \(F_{\max}\) & \(F_{\mathrm{wmax}}\) \\
\midrule
\(K=8\) & \(0.611430\pm0.011644\) & \(0.509754\pm0.019627\) \\
\(K=16\) & \(0.623143\pm0.004861\) & \(0.527054\pm0.011763\) \\
\bottomrule
\end{tabular}
\end{table}

\paragraph{Alternative sampling.}
Diversity-based selection yields \(F_{\max}/F_{\mathrm{wmax}}=0.62200/0.52444\) for \(K=8\) and \(0.61336/0.51258\) for \(K=16\), compared with \(0.62267/0.52527\) for all 32 conformers (Table~\ref{tab:diversity_sampling}). These results show comparable performance under an alternative conformer-selection strategy.

\begin{table}[htbp]
\centering
\small
\caption{Diversity-based subset selection compared with the full ensemble.}
\label{tab:diversity_sampling}
\begin{tabular}{lrr}
\toprule
Selection & \(F_{\max}\) & \(F_{\mathrm{wmax}}\) \\
\midrule
Diversity-based \(K=8\) & 0.62200 & 0.52444 \\
Diversity-based \(K=16\) & 0.61336 & 0.51258 \\
Full \(K=32\) reference & 0.62267 & 0.52527 \\
\bottomrule
\end{tabular}
\end{table}

\paragraph{Permutation robustness.}
Shuffling the order of all 32 conformers leaves the aggregate metrics unchanged at \(F_{\max}=0.62267\) and \(F_{\mathrm{wmax}}=0.52527\). Across 118 records and 166 protein--aspect examples, no predicted GO-term set changes. This is consistent with the intended permutation-invariant aggregation of an unordered conformer set.

\section{Ensemble Negative Controls}
\label{app:negative}

\paragraph{Protein--ensemble correspondence.}
A mismatched-ensemble control breaks the correspondence between a protein and its structural ensemble. The ensemble reference model obtains \(F_{\max}/F_{\mathrm{wmax}}=0.62267/0.52527\), whereas mismatched ensembles yield \(0.42431/0.25065\). The paired full-minus-mismatch differences are \(+0.198367\) (95\% CI \([+0.149577,+0.247620]\)) for \(F_{\max}\) and \(+0.274623\) (95\% CI \([+0.203303,+0.347978]\)) for \(F_{\mathrm{wmax}}\). Correct protein-specific ensemble information therefore contributes substantially to prediction.

\paragraph{Distinct conformers versus a repeated single state.}
The repeat-single control replaces the diverse conformer input with repeated copies of one conformer. All five repeat-single evaluations fall below the full ensemble reference on both primary metrics (Table~\ref{tab:repeat_single}).

\begin{table}[htbp]
\centering
\small
\caption{Repeat-single negative control. Each row supplies repeated copies of one structural state; the full reference uses 32 distinct conformers.}
\label{tab:repeat_single}
\begin{tabular}{lrr}
\toprule
Input & \(F_{\max}\) & \(F_{\mathrm{wmax}}\) \\
\midrule
Repeated single conformer 1 & 0.50902 & 0.39536 \\
Repeated single conformer 2 & 0.55108 & 0.44531 \\
Repeated single conformer 3 & 0.51110 & 0.38598 \\
Repeated single conformer 4 & 0.53560 & 0.42086 \\
Repeated single conformer 5 & 0.55238 & 0.44521 \\
Full ensemble reference & \textbf{0.62267} & \textbf{0.52527} \\
\bottomrule
\end{tabular}
\end{table}

\section{Variability Adjustment and Covariate Matching}
\label{app:variability}

The original fixed-quartile analysis divides the 118 test records into groups of 30, 30, 29, and 29 using the mean pairwise C\(\alpha\) TM-score of their 32 conformers. Lower within-ensemble TM-score denotes greater conformational variability. The adjusted analyses consider training ESM2 sequence similarity, log protein length, propagated annotation count, and training-derived GO rarity/frequency as covariates.

\paragraph{Adjusted high-variability contrast.}
Relative to ESM2 sequence 5NN, the adjusted Q4 gain is \(0.183189\), compared with \(-0.108557\) for the remaining records. The Q4-minus-rest contrast is \(+0.291746\), with protein-bootstrap 95\% CI \([0.095152,0.487881]\) and related-group-bootstrap 95\% CI \([0.094825,0.492712]\). The elevated benefit in the highest-variability quartile remains after adjustment for the four covariate groups.

\paragraph{Continuous embedding-variability analysis.}
The adjusted gain difference between the 90th and 50th embedding-variability percentiles is \(+0.134365\), with 95\% CI \([0.006773,0.252423]\). The continuous adjusted curve is consistent with an increase in ensemble benefit toward the high-variability regime followed by saturation.

\paragraph{Matching robustness.}
Covariate matching produces 13 pairs. The mean within-pair gain difference relative to ESM2 sequence 5NN is \(+0.501261\); 11 pairs are positive, one is negative, and one is zero. Matching therefore provides supporting evidence for a larger ensemble benefit among high-variability records.

\begin{table}[htbp]
\centering
\small
\caption{Selected covariate-adjusted variability contrasts relative to ESM2 sequence 5NN.}
\label{tab:adjusted_variability}
\begin{tabular}{lrr}
\toprule
Contrast & Estimate & 95\% CI \\
\midrule
Q4 minus rest, protein bootstrap & 0.291746 & [0.095152, 0.487881] \\
Q4 minus rest, related-group bootstrap & 0.291746 & [0.094825, 0.492712] \\
90th minus 50th embedding-variability percentile & 0.134365 & [0.006773, 0.252423] \\
Matched high-variability gain difference (13 pairs) & 0.501261 & --- \\
\bottomrule
\end{tabular}
\end{table}

\section{Quantitative Evaluation of Functional Reasoning}
\label{app:reasoning}

\paragraph{Paired evaluation design.}
The evaluation uses 236 responses comprising 118 matched sequence-GO/ensemble-GO response pairs. Only the GO-anchor condition differs between conditions; sequence, organism information, InterPro preprocessing, prompting, the reasoning model, and decoding settings are fixed. A blinded Qwen2.5-32B-Instruct judge evaluates generated biological statements against the available reference evidence. Fractions are computed over checkable units within each response and averaged across proteins. Confidence intervals use paired protein bootstrap with 10,000 replicates.

\paragraph{Reference grounding.}
Reference-supported content increases under ensemble-conditioned anchors (Table~\ref{tab:reasoning_grounding}). The reference-supported fraction increases by \(1.8406\) percentage points (95\% CI \([1.0029,2.6816]\)). The reference-unverifiable fraction decreases by \(2.0520\) percentage points (95\% CI \([1.1842,2.9159]\)). Reference-unverifiable identifies statements that the available reference evidence does not adjudicate. There is no statistically detectable increase in unsupported-specific assertions and no statistically detectable difference in reference contradictions.

\begin{table}[t]
\centering
\caption{\textbf{Ensemble-conditioned GO anchors increase reference-supported content.} Differences are ensemble minus sequence. Fractions are response-level means; pp denotes percentage points. Intervals are paired protein-bootstrap 95\% CIs.}
\label{tab:reasoning_grounding}
\begin{tabular}{@{}lrl@{}}
\toprule
Measure & Paired change & 95\% CI\\
\midrule
Reference-supported fraction & +1.8406 pp & [1.0029, 2.6816]\\
Reference-unverifiable fraction & $-2.0520$ pp & [$-2.9159$, $-1.1842$]\\
Supported units per response & +0.423729 & [0.220339, 0.627119]\\
\bottomrule
\end{tabular}
\end{table}

\paragraph{Absolute content and length controls.}
Mean absolute reference-supported units per response increase from \(0.483051\) to \(0.906780\), a paired difference of \(+0.423729\) (95\% CI \([+0.220339,+0.627119]\)). Controlling for checkable-unit count yields a support-rate ratio of \(1.970460\) (95\% CI \([1.433944,2.840275]\)); controlling for response word count yields \(1.947284\) (95\% CI \([1.405645,2.819655]\)).

\begin{table}[htbp]
\centering
\small
\caption{Reference-supported content after controlling for response size. Ratios compare ensemble-GO with sequence-GO responses.}
\label{tab:reasoning_length_controls}
\begin{tabular}{lrr}
\toprule
Control & Support-rate ratio & 95\% CI \\
\midrule
Checkable-unit count & 1.970460 & [1.433944, 2.840275] \\
Response word count & 1.947284 & [1.405645, 2.819655] \\
\bottomrule
\end{tabular}
\end{table}

\section{Biological Process Prediction Coverage}
\label{app:bp}

Biological Process has denser propagated truth annotations than Cellular Component, while the number of returned positive predictions is similar (Table~\ref{tab:bp_coverage}). For BP, 31 eligible examples have on average 17.48 propagated truth labels and 4.74 returned positive predictions, with recall \(0.2190\), IA recall \(0.1910\), and AP \(0.1970\). This describes a prediction-coverage gap that contributes to the weaker BP performance.

\begin{table}[htbp]
\centering
\small
\caption{Aspect-specific prediction coverage. Truth labels and returned predictions are mean counts per eligible example; recall, IA recall, and AP summarize prediction coverage and ranking.}
\label{tab:bp_coverage}
\begin{tabular}{lrrrrrr}
\toprule
Aspect & \(n\) & Truth labels & Returned & Recall & IA recall & AP \\
\midrule
BP & 31 & 17.48 & 4.74 & 0.2190 & 0.1910 & 0.1970 \\
CC & 107 & 6.84 & 4.00 & 0.6790 & 0.5730 & 0.5820 \\
MF & 28 & 8.61 & 5.68 & 0.4160 & 0.3190 & 0.3060 \\
\bottomrule
\end{tabular}
\end{table}

\section{Multi-Seed Training Variance}
\label{app:seeds}

The matched-budget evaluation comprises three independently initialized \(K=32\) runs. The runs yield test \(F_{\max}=0.613200\pm0.007543\) and \(F_{\mathrm{wmax}}=0.528933\pm0.012570\), reported as mean \(\pm\) sample standard deviation over three runs. The independent runs with seeds 20261001 and 20261002 both reach or exceed the main benchmark result of \(F_{\max}/F_{\mathrm{wmax}}=0.6045/0.5172\) (seed 20260425), as shown in Table~\ref{tab:seed_reproduction}.

\begin{table}[htbp]
\centering
\small
\caption{Training variance and reproduction of the original benchmark result. The three-run row summarizes the matched-budget evaluation.}
\label{tab:seed_reproduction}
\begin{tabular}{llrr}
\toprule
Model / summary & Training seed & \(F_{\max}\) & \(F_{\mathrm{wmax}}\) \\
\midrule
Main benchmark model & 20260425 & 0.6045 & 0.5172 \\
Independent initialization & 20261001 & 0.6179 & 0.5422 \\
Independent initialization & 20261002 & 0.6172 & 0.5274 \\
Three runs, mean \(\pm\) SD & --- & \(0.613200\pm0.007543\) & \(0.528933\pm0.012570\) \\
\bottomrule
\end{tabular}
\end{table}

    \FloatBarrier

    \section{Evaluation Metrics}
    \label{app:metrics}
    
    All metrics are computed on propagated non-root GO terms. For evaluation, each example \(i\) denotes a protein-aspect pair, where the aspect is one of MF, BP, or CC. Ground-truth and predicted GO terms are propagated within the corresponding GO aspect, excluding the aspect root. Let \(T_i\) denote the propagated ground-truth GO set for example \(i\), and let \(s_i(g)\) be the predicted score for GO term \(g\). Unreported GO terms are treated as zero-score terms and therefore do not enter positive-threshold predictions. Let \(\mathrm{IA}(g)\) denote the information accretion weight of term \(g\).
    
    \paragraph{CAFA-style metrics.}
    We use CAFA-style metrics to denote threshold-swept, precision-recall, and semantic-distance metrics commonly used in GO-based protein function prediction evaluation~\citep{radivojac2013large,clark2013information,kulmanov2024semantic}. For a score threshold \(\tau\), the predicted set for example \(i\) is
    \[
    P_i(\tau)=\{g:s_i(g)\geq \tau\}.
    \]
    For unweighted \(F_{\max}\), the per-example precision and recall are
    \[
    \mathrm{Prec}_i(\tau)
    =
    \frac{|P_i(\tau)\cap T_i|}{|P_i(\tau)|},
    \qquad
    \mathrm{Rec}_i(\tau)
    =
    \frac{|P_i(\tau)\cap T_i|}{|T_i|}.
    \]
    Following the CAFA-style convention used in our evaluation, precision is averaged only over examples with non-empty predictions, while recall is averaged over all examples:
    \[
    \mathrm{Prec}(\tau)
    =
    \frac{1}{|\mathcal{I}_{\tau}^{+}|}
    \sum_{i\in \mathcal{I}_{\tau}^{+}}
    \mathrm{Prec}_i(\tau),
    \qquad
    \mathrm{Rec}(\tau)
    =
    \frac{1}{N}
    \sum_{i=1}^{N}
    \mathrm{Rec}_i(\tau),
    \]
    where
    \[
    \mathcal{I}_{\tau}^{+}
    =
    \{i:|P_i(\tau)|>0\}.
    \]
    The threshold-specific F-score is
    \[
    F(\tau)
    =
    \frac{
    2\,\mathrm{Prec}(\tau)\,\mathrm{Rec}(\tau)
    }{
    \mathrm{Prec}(\tau)+\mathrm{Rec}(\tau)
    },
    \]
    and \(F_{\max}\) is
    \[
    F_{\max}
    =
    \max_{\tau} F(\tau).
    \]
    
    \(F_{\mathrm{wmax}}\) denotes weighted \(F_{\max}\), i.e., the IA-weighted version of \(F_{\max}\). For each example, we replace unweighted counts with sums of IA weights:
    \[
    \mathrm{Prec}^{w}_i(\tau)
    =
    \frac{
    \sum_{g\in P_i(\tau)\cap T_i}\mathrm{IA}(g)
    }{
    \sum_{g\in P_i(\tau)}\mathrm{IA}(g)
    },
    \qquad
    \mathrm{Rec}^{w}_i(\tau)
    =
    \frac{
    \sum_{g\in P_i(\tau)\cap T_i}\mathrm{IA}(g)
    }{
    \sum_{g\in T_i}\mathrm{IA}(g)
    }.
    \]
    Weighted precision and recall are then averaged in the same way as above:
    \[
    \mathrm{Prec}^{w}(\tau)
    =
    \frac{1}{|\mathcal{I}_{\tau}^{+}|}
    \sum_{i\in \mathcal{I}_{\tau}^{+}}
    \mathrm{Prec}^{w}_i(\tau),
    \qquad
    \mathrm{Rec}^{w}(\tau)
    =
    \frac{1}{N}
    \sum_{i=1}^{N}
    \mathrm{Rec}^{w}_i(\tau).
    \]
    The weighted F-score is
    \[
    F^{w}(\tau)
    =
    \frac{
    2\,\mathrm{Prec}^{w}(\tau)\,\mathrm{Rec}^{w}(\tau)
    }{
    \mathrm{Prec}^{w}(\tau)+\mathrm{Rec}^{w}(\tau)
    },
    \]
    and
    \[
    F_{\mathrm{wmax}}
    =
    \max_{\tau}F^{w}(\tau).
    \]
    We report overall \(F_{\max}/F_{\mathrm{wmax}}\), aspect-specific MF/BP/CC \(F_{\max}/F_{\mathrm{wmax}}\), high-IA \(F_{\max}/F_{\mathrm{wmax}}\), AUPR, and \(S_{\min}\). AUPR is the area under the threshold-swept precision--recall curve defined above. For semantic distance, remaining uncertainty and misinformation at threshold \(\tau\) are
    \[
    \mathrm{ru}(\tau)
    =
    \frac{1}{N}\sum_{i=1}^{N}\sum_{g\in T_i\setminus P_i(\tau)}\mathrm{IA}(g),
    \qquad
    \mathrm{mi}(\tau)
    =
    \frac{1}{N}\sum_{i=1}^{N}\sum_{g\in P_i(\tau)\setminus T_i}\mathrm{IA}(g),
    \]
    and
    \[
    S_{\min}
    =
    \min_{\tau}\sqrt{\mathrm{ru}(\tau)^2+\mathrm{mi}(\tau)^2}.
    \]
    Lower \(S_{\min}\) indicates smaller semantic error. The high-IA subset is defined by the 75th percentile of IA among propagated ground-truth test terms, corresponding to approximately \(\mathrm{IA}\geq 1.684844\).
    
    \paragraph{Protein-aspect ranking setup.}
    For ranking metrics, each protein-aspect example defines one ranking problem. Let
    \[
    (g_{i,1},g_{i,2},\ldots)
    \]
    be the positive-score propagated predicted GO terms sorted by decreasing score \(s_i(g)\). Let
    \[
    \mathrm{Pred}_{i,k}
    =
    \{g_{i,1},\ldots,g_{i,\min(k,M_i)}\}
    \]
    denote the top up to \(k\) positive-score predictions, where \(M_i\) is the number of positive-score propagated predictions for example \(i\). Thus, if fewer than \(k\) positive-score terms are available, \(\mathrm{Pred}_{i,k}\) contains fewer than \(k\) terms. Metrics are computed per example and then averaged over protein-aspect examples.
    
    \paragraph{IA-MAP.}
    IA-MAP is an information-weighted protein-aspect mean average precision. Let \(a(g)=\mathrm{IA}(g)\). For example \(i\), define the IA-weighted precision at rank \(j\) as
    \[
    P_i^{\mathrm{IA}}(j)
    =
    \frac{
    \sum_{k=1}^{j}
    a(g_{i,k})\,
    \mathbf{1}[g_{i,k}\in T_i]
    }{
    \sum_{k=1}^{j}a(g_{i,k})
    }.
    \]
    The incremental IA-weighted recall at rank \(j\) is
    \[
    \Delta R_i^{\mathrm{IA}}(j)
    =
    \frac{
    a(g_{i,j})\,
    \mathbf{1}[g_{i,j}\in T_i]
    }{
    \sum_{g\in T_i}a(g)
    }.
    \]
    The IA-weighted average precision for example \(i\) and its mean over all examples are
    \[
    \mathrm{IA\mbox{-}AP}_i
    =
    \sum_{j=1}^{M_i}
    P_i^{\mathrm{IA}}(j)
    \Delta R_i^{\mathrm{IA}}(j),
    \qquad
    \mathrm{IA\mbox{-}MAP}
    =
    \frac{1}{N}
    \sum_{i=1}^{N}
    \mathrm{IA\mbox{-}AP}_i.
    \]
    A ground-truth GO term contributes more when it appears earlier in the ranked list, and higher-IA ground-truth terms contribute more strongly.
    
    \paragraph{NDCG@\(k\).}
    NDCG@\(k\) evaluates top-\(k\) ranking quality using IA as graded relevance~\citep{jarvelin2002cumulated}. For example \(i\), the discounted cumulative gain is
    \[
    \mathrm{DCG@}k_i
    =
    \sum_{j=1}^{\min(k,M_i)}
    \frac{
    \mathbf{1}[g_{i,j}\in T_i]\,\mathrm{IA}(g_{i,j})
    }{
    \log_2(j+1)
    }.
    \]
    Let \(\mathrm{IDCG@}k_i\) be the maximum possible DCG@\(k\), obtained by ranking ground-truth terms by decreasing IA. Then
    \[
    \mathrm{NDCG@}k_i
    =
    \frac{\mathrm{DCG@}k_i}{\mathrm{IDCG@}k_i},
    \qquad
    \mathrm{NDCG@}k
    =
    \frac{1}{N}
    \sum_{i=1}^{N}
    \mathrm{NDCG@}k_i.
    \]
    We report NDCG@1, NDCG@3, NDCG@5, and NDCG@10.
    
    \paragraph{Jaccard@\(k\).}
    Jaccard@\(k\) compares the top up to \(k\) positive-score predicted GO set with the ground-truth GO set:
    \[
    \mathrm{Jaccard@}k_i
    =
    \frac{
    |\mathrm{Pred}_{i,k}\cap T_i|
    }{
    |\mathrm{Pred}_{i,k}\cup T_i|
    },
    \qquad
    \mathrm{Jaccard@}k
    =
    \frac{1}{N}
    \sum_{i=1}^{N}
    \mathrm{Jaccard@}k_i.
    \]
    We report Jaccard@3, Jaccard@5, and Jaccard@10.
    
    \paragraph{IA-R@\(k\).}
    IA-R@\(k\) measures the fraction of ground-truth GO information recovered in the top up to \(k\) positive-score predictions:
    \[
    \mathrm{IA\mbox{-}R@}k_i
    =
    \frac{
    \sum_{g\in \mathrm{Pred}_{i,k}\cap T_i}\mathrm{IA}(g)
    }{
    \sum_{g\in T_i}\mathrm{IA}(g)
    },
    \qquad
    \mathrm{IA\mbox{-}R@}k
    =
    \frac{1}{N}
    \sum_{i=1}^{N}
    \mathrm{IA\mbox{-}R@}k_i.
    \]
    We report IA-R@1, IA-R@3, and IA-R@5.
    
    \paragraph{CC-specific ranking metrics.}
    We additionally report CC IA-MAP and CC NDCG@1/3/5/10, computed only within the Cellular Component branch. Cellular Component annotations describe where a gene product acts, including cellular structures, subcellular compartments, and stable protein-containing complexes. These labels are therefore naturally closer to structural context, localization, and molecular organization than many process-level annotations. This makes CC a particularly informative aspect for evaluating whether residue-aligned conformational ensemble conditioning improves GO prediction.
    
    \section{Baseline Details}
    \label{app:baselines}
    
    \paragraph{DeepGO.}
    We evaluate DeepGO~\citep{kulmanov2018deepgo} using the released aspect-specific MF, BP, and CC checkpoints. Query sequences are converted to n-gram representations, truncated to 1,000 residue positions, and searched with DIAMOND against the released DeepGO database~\citep{buchfink2015diamond}. The retrieved 256-dimensional graph embedding and sequence representation are processed jointly by each aspect-specific model. We retain continuous scores and evaluate recognized GO identifiers with the common propagation and evaluation pipeline.
    
    \paragraph{DeepFRI-CNN.}
    DeepFRI-CNN~\citep{gligorijevic2021structure} is a sequence-based external baseline. It predicts GO annotations from sequence-derived representations without using PED conformational ensembles. We evaluate its output on the same test proteins and metrics.
    
    \paragraph{DeepFRI-GCN.}
    DeepFRI-GCN~\citep{gligorijevic2021structure} is a structure-based external baseline. For each test protein, we use the matched AFDB single-structure model as the structure input. This baseline tests the performance of a pretrained structure-based GO predictor under a single-structure setting.

    \paragraph{BioReason-Pro.}
    BioReason-Pro~\citep{fallahpour2026bioreason} is the LLM-based baseline. It combines sequence-only ontology predictions with the protein sequence, organism information, and InterPro annotations to generate structured functional outputs. For evaluation, generated GO identifiers are parsed and assigned a score of 1.0 under the common propagation and evaluation pipeline; the fixed reasoning-module comparison is detailed in Appendix~\ref{app:implementation}.
    
    \paragraph{GO prior.}
    The GO-prior row in the main tables is the aspect-frequency prior. We count term frequencies within MF, BP, and CC from the fixed training split and normalize each aspect by its maximum term frequency. The resulting aspect-specific score vector is applied to every test protein and evaluated with the common hierarchy propagation and root-exclusion procedure.
    
    \paragraph{Length nearest-neighbor transfer.}
    Length 5NN identifies the five training proteins with the closest sequence lengths and accumulates transferred GO scores using \(s(x_i,x_j)=1/(1+|L_i-L_j|)\). This baseline tests the information provided by sequence length alone.
    
    \paragraph{Amino-acid composition transfer.}
    AA composition 5NN represents each protein by a 20-dimensional amino-acid composition vector and uses cosine similarity for nearest-neighbor label transfer. It captures global sequence composition while ignoring residue order.
    
    \paragraph{ESM2 sequence retrieval.}
    We encode sequences with \texttt{facebook/esm2\_t36\_3B\_UR50D} and use hidden states from layer 30~\citep{lin2023evolutionary}. Sequences are uppercased, truncated to 1,024 residues, and tokenized without special tokens. Valid residue representations are mean-pooled and \(L_2\)-normalized to form one embedding per protein. The neighbor pool contains the 5,654 labeled proteins in the fixed training split; validation and test proteins are excluded. The 1NN baseline transfers labels from the highest-cosine training protein. For 5NN, each term receives \(u(g)=\sum_{j=1}^{5}\max(\cos(q,n_j),0)\mathbf{1}[g\in Y_{n_j}]\), followed by within-protein normalization \(s(g)=u(g)/\max_h u(h)\). Transferred scores are processed by the common GO propagation and evaluation pipeline.
    
    \paragraph{ESM3 AFDB structure retrieval.}
    We generate one embedding per protein with the open ESM3 small checkpoint, \texttt{esm3\_sm\_open\_v1}, using the sequence and matched AFDB coordinates~\citep{hayes2025simulating}. Residue representations are mean-pooled and \(L_2\)-normalized. These embeddings are AFDB structure-conditioned representations, rather than structure-only or PED ensemble embeddings. The 1NN and 5NN retrieval rules, training pool, score normalization, and leakage controls match the ESM2 baselines. AFDB embeddings are available for all 118 test proteins.
    
    \section{Implementation Details and Hyperparameters}
    \label{app:implementation}
    
    \paragraph{Backbone and frozen components.}
    The ontology predictor is initialized from a pretrained sequence-only autoregressive model. The sequence encoder is ESM2~\citep{lin2023evolutionary}, and we use sequence hidden layer 30. The autoregressive decoder has 12 layers, hidden size 900, and 12 attention heads. The GO vocabulary size is 10,303, the organism vocabulary size is 201, and the context block size is 2,048. During ensemble-conditioning training, we freeze the ESM2 sequence encoder, protein projection, GO autoregressive decoder, GO-token embeddings, organism embeddings, and output head.
    
    \paragraph{Trainable ensemble-conditioning components.}
    The trainable modules are the residue-wise ensemble encoder and the learned sequence-prefix and structure-prefix type embeddings. We use the \texttt{ensemble\_residue} structure mode. The ESM3 feature dimension is 1,536~\citep{hayes2025simulating}, and each protein uses \(K=32\) conformers. The residue ensemble encoder hidden size is 900 and uses 2 set-attention layers. The number of attention heads in the ensemble encoder is 12, matched to the autoregressive decoder when \texttt{ensemble\_num\_heads=None}. Each set-attention block uses pre-layer-normalized multi-head self-attention with a residual connection, followed by a pre-layer-normalized MLP/FFN with another residual connection~\citep{vaswani2017attention,lee2019set,xiong2020layer}. The MLP expansion ratio is \(4\times\), the activation is GELU~\citep{hendrycks2016gelu}, and dropout is 0.1. The conformer set at each residue is pooled by learned-query multi-head attention, followed by output LayerNorm. Learned sequence-prefix and structure-prefix type embeddings are added before concatenation.
    
    \paragraph{Input and length settings.}
    The maximum requested protein length is 1,024, the reserved GO-token budget is 256, and the effective maximum protein length in residue-structure mode is 896. This is because the sequence prefix and residue-aligned structure prefix are concatenated, giving approximately \(2L\) prefix tokens under the 2,048 block-size budget. The protein context is
    \[
    [
    \text{sequence prefix}+\text{sequence type embedding};
    \text{structure prefix}+\text{structure type embedding}
    ].
    \]
    
    \paragraph{Optimization.}
    Training uses the autoregressive GO-token negative log-likelihood with teacher forcing~\citep{williams1989learning}. We use AdamW~\citep{loshchilov2019decoupled} with learning rate \(10^{-4}\), weight decay \(10^{-2}\), batch size 2, and gradient accumulation over 4 steps, giving an effective batch size of 8 aspect-level training examples. We use gradient clipping with norm 1.0 and bf16 mixed precision.
    
    \paragraph{Training data accounting.}
    The protein split is the PED train/validation/test split described in the main evaluation setup. The ensemble training loader contains 15,191 aspect-level GO-sequence examples, and the validation loader contains 190 aspect-level GO-sequence examples. The benchmark test split contains 118 PED entry/ensemble records. Analyses that require aspect-specific ground truth use the corresponding eligible protein-aspect examples.
    
    \paragraph{Reasoning-module instantiation.}
We hold the InterPro preprocessing, prompting, RL checkpoint, and decoding settings of the LLM reasoning module fixed. The controlled comparison changes only its ontology anchors: the baseline uses predictions from the sequence-only ontology predictor, whereas DynaReason-Pro supplies ensemble-conditioned predictions. For evaluation, we parse generated outputs to extract GO identifiers and assign each extracted term a score of 1.0. The paired reference-grounding evaluation is detailed in Appendix~\ref{app:reasoning}.

\end{document}